\documentclass[12pt]{article}
\usepackage{a4,graphicx,fancyhdr, vmargin, psfig, texstyle, comment, booktabs, appendix}
\usepackage[style=apa,backend=biber]{biblatex}
\begin{document}

\begin{flushleft}
  {\LARGE\bf Generative Adversarial Networks for\\[2mm]
  Synthetic Data Generation: A Comparative Study} 
  \vspace{1.0cm}
  
  Claire Little$^*$, Mark Elliot$^*$ Richard Allmendinger$^{**}$ Sahel Shariati Samani$^*$

{\small
\begin{description}
\item $^* \;$ School of Social Sciences, University of Manchester, Manchester,
M13 9PL, UK
\item $^{**}\;$ Alliance Manchester Business School,
University of Manchester, Manchester,
M13 9PL, UK
\end{description}
}
\end{flushleft}
\vspace{1cm}

\noindent {\bf Abstract}.
{\small Generative Adversarial Networks (GANs) are gaining increasing attention as a means for synthesising data. So far much of this work has been applied to use cases outside of the data confidentiality domain with a common application being the production of artificial images. Here we consider the potential application of GANs for the purpose of generating synthetic census microdata. We employ a battery of utility metrics and a disclosure risk metric (the Targeted Correct Attribution Probability) to compare the data produced by tabular GANs with those produced using orthodox data synthesis methods. }


\section{Introduction}

Machine Learning (ML) methods are showing increasing promise as an approach to synthetic data generation. Generative Adversarial Networks (GANs), first proposed by~\textcite{Goodfellow2014GenerativeNets}, are the focus of much of the research literature. GANs are a generative deep learning technique that use artificial neural networks. The generative element of a GAN does not access the original data whilst training, and can therefore produce synthetic data without directly interacting with the original data; this is an interesting feature that might in theory reduce disclosure risk.

Whilst research into using GANs for mixed-type, or tabular, microdata has so far been limited, GANs are generating much research interest and have been used for various applications, although as detailed by~\textcite{Wang2020ANetworks} these are predominantly in the image domain. Research has focussed on purposes such as: synthetic image generation (e.g.~\textcite{Karras2019ANetworks, Radford2016UnsupervisedNetworks}); image colorization (e.g.~\textcite{Nazeri2018ImageNetworks}); image-to-image translation (e.g.~\textcite{Huang2018MultimodalTranslation, Isola2017Image-to-imageNetworks, Zhu2017UnpairedNetworks}); image inpainting (e.g.~\textcite{Demir2018Patch-BasedNetworks}); super-resolution (e.g.~\textcite{Menon2020PULSE:Models, Ledig2017Photo-realisticNetwork}); and synthetic medical image generation (e.g.~\textcite{Piacentino2021GeneratingData, Sandfort2019DataTasks, Frid-Adar2018GAN-basedClassification, Iqbal2018GenerativeMI-GAN}).   

In this paper, we provide a framework for assessing the relative merits of GANs compared to traditional statistical methods for producing synthetic data in terms of the utility and residual risk of the data produced. We first give a brief introduction to the data synthesis problem and deep learning approaches in Section \ref{sec:Background}. In section \ref{sec:Design} we outline the design of study. Section \ref{sec:Results} provides the results of comparing two GANS with CART using \textit{Synthpop} and a Bayesian approach using \textit{DataSynthesizer}.  

\section{Background}
\label{sec:Background}
\subsection{Data Synthesis}

\textcite{Rubin1993StatisticalLimitation} introduced the idea of synthetic data, proposing using multiple imputation on all variables such that none of the original data was released. \textcite{Little1993StatisticalData} proposed an alternative that simulated only sensitive variables, thereby producing partially synthetic data. Rubin's idea was slow to be adopted, as noted by~\textcite{Raghunathan2003MultipleLimitation}, who along with~\textcite{Reiter2002SatisfyingSets, Reiter2003InferenceSets, Reiter2003ReleasingStudy}, formalised the synthetic data problem. Further work has involved using non-parametric methods such as classification and regression trees (CART) and random forests (e.g. ~\textcite{Reiter2005UsingMicrodata, Drechsler2010SamplingMicrodata, Drechsler2011AnDatasets})

There are two competing objectives when producing synthetic data: high data utility (i.e., ensuring that the synthetic data is useful, with a distribution close to the original) and low disclosure risk. Balancing this trade-off can be difficult, as, in general, reducing disclosure risk comes at a cost in utility. This trade-off can be visualised by considering the R-U confidentiality map developed by~\textcite{Duncan2004DatabaseMap}. Whilst there are multiple measures of utility, ranging from comparing summary statistics, correlations and cross-tabulations, to considering data performance using predictive algorithms, there are fewer that focus on disclosure risk for synthetic data. As noted by~\textcite{Taub2018DifferentialExploration}, much of the statistical disclosure control (SDC) literature focusses on re-identification risk, which is not meaningful for synthetic data, rather than the risk of attribution, which is more likely. The Targeted Correct Attribution Probability (TCAP) developed by~\textcite{Elliot2014FinalTeam, Taub2018DifferentialExploration} can be used to assess attribution risk.

\subsection{Deep Learning and GANs}

Deep learning~\parencite{Lecun2015DeepLearning}, a subset of the broader field of machine learning, uses artificial neural networks to learn models from data. Neural networks (NNs) are made up of a series of stacked layers of neurons joined by weighted connections (the term ‘deep’ refers to the number of hidden layers; a ‘shallow’ NN may contain only 1 or 2 layers). In general, a NN is trained and learns iteratively by backpropagating the loss, or error, through the network and adjusting the weights to reach an optimal solution. Deep learning methods can discover the underlying structure in complex, high-dimensional data and have been responsible for dramatically improved performance in areas such as speech recognition, image recognition, object detection, natural language understanding and genomics~\parencite{Lecun2015DeepLearning}.

GANs~\parencite{Goodfellow2014GenerativeNets}, simultaneously train two NN models: a generative model which captures the data distribution, and a discriminative model that aims to determine whether a sample is from the model distribution or the data distribution. The process corresponds to a minimax two-player game. The generative model starts off with noise as inputs (it does not access the training, or original, dataset at all) and relies on feedback from the discriminative model to generate a data sample. 

\begin{figure}
\centering
\includegraphics[scale=0.5]{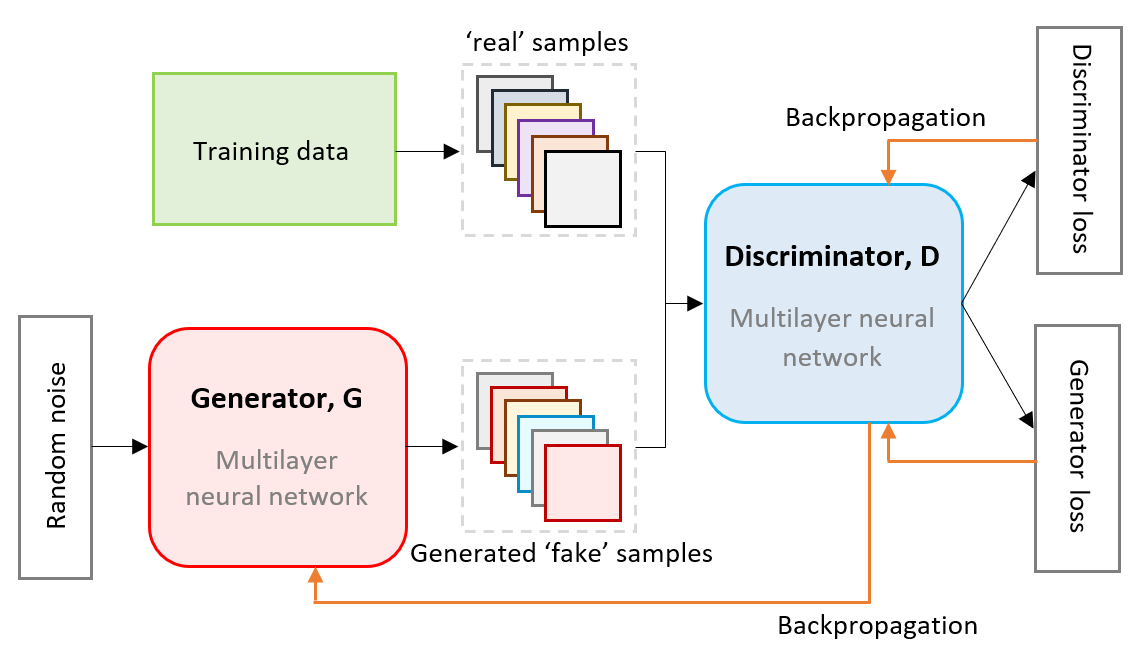}
\setlength{\abovecaptionskip}{-5pt}
\setlength{\belowcaptionskip}{-65pt}
\caption{\label{fig:GAN2}Example of GAN Architecture}
\end{figure}

As described by~\textcite{Goodfellow2014GenerativeNets}, the discriminative and generative models are typically both multilayer NNs that are trained using the backpropagation or dropout algorithms. GANs perform alternate training, whereby the discriminator trains whilst the generator is held constant, and vice versa. The discriminator can be thought of as a supervised classification model. It receives batches of labelled real and generated data examples and outputs a single value for each example, the probability that it came from the real distribution, rather than the generator. If this value is close to 1, then it would be considered real; closer to zero would be classified as fake. The discriminator is penalized for misclassifying fake/real instances, and the weights adjusted accordingly. During generator training the weights are updated based on how well the generated samples fool the discriminator (ideally, when a generated image is fed into the discriminator the output will be close to 1). Figure \ref{fig:GAN2} contains a basic example of GAN architecture.

GAN training can be challenging to optimize, as it can be difficult to balance the training of both models (generator and discriminator). If they do not learn at a similar rate then the feedback may not be useful. GANs can also be susceptible to issues such as vanishing gradients (where the discriminator does not feedback enough information for the generator to learn), mode collapse (e.g., the generator finds a small number of samples that fool the discriminator and only produces those, leading to the gradient of the loss function to collapse to near 0) and failure to converge. And, as noted by~\textcite{Lucic2018AreGansCreatedEqual}, since there is no consistent and generally accepted evaluation metric, it can be difficult to objectively evaluate or compare the performance of different models.

Microdata tends to contain mixed data (i.e., a combination of numerical and categorical variables), which is likely to be heterogeneous, containing imbalanced categorical variables, and skewed or multimodal numerical distributions. GANs for image generation tend to deal with numerical, homogeneous data; in general, they must be adapted to deal with mixed data. Several studies have done this by adapting the GAN architecture, these are often referred to as tabular GANs. medGAN, developed by~\textcite{Choi2017GeneratingNetworks} combined an autoencoder with a GAN to produce synthetic electronic health record (EHR) data containing binary (but not multi categorical) and continuous data. \textcite{Camino2018GeneratingNetworks} extended this work to include categorical data, however experiments by~\textcite{Goncalves2020GenerationData} found the model failed to generate realistic patient data. \textcite{Chen2019Faketables:Data} proposed ITS-GAN which used a convolutional GAN architecture (normally used for images) and autoencoders to encode the “functional dependencies” within the data.

TableGAN, developed by~\textcite{Park2018DataNetworks} is based on the convolutional DCGAN~\parencite{Radford2016UnsupervisedNetworks} architecture, but contains three NNs (generator, discriminator and classifier) as opposed to the standard two NNs. The classifier NN is used to learn the “semantics” or rules from the original data and incorporate them into the training process. CTAB-GAN by~\textcite{Zhao2021CTAB-GAN:Synthesizing} is based on a conditional GAN and also incorporates a classifier which is designed to learn the semantics of the data. TGAN, proposed by~\textcite{Xu2018SynthesizingNetworks} uses a Recurrent Neural Network (RNN) architecture with LSTM (long short-term memory) cells for the generator. RNNs are often used to process sequences of data (such as speech), and using this architecture, TGAN produces data column by column, predicting the value for the next column based on the previous ones. CTGAN, developed by~\textcite{Xu2019ModelingGAN} is by the authors of TGAN, but does not use the same RNN GAN architecture. CTGAN uses "mode-specific normalization" to overcome non-Gaussian and multimodal distribution problems, and employs oversampling methods to handle class imbalance in the categorical variables. Much as with the GANs designed for numeric data, it is notable across much of the tabular GAN research, that there is inconsistency in terms of how they are evaluated, with no dominant method employed (other than assessing machine learning utility).

\section{Study design} 
\label{sec:Design}
This study compares the performance of tabular GANs with more orthodox data synthesis methods.

\subsection{Data}
The 1991 Individual Sample of Anonymised Records (SAR) for the British Census~\parencite{OfficeforNationalStatisticsCensusDivisionUniversityofManchesterCathieMarshCentreforCensusandSurveyResearch2013CensusSARs} was used. This contains a 2\% sample (1,116,181 records) of the population of Great Britain (excluding Northern Ireland), including adults and children. There are 67 variables containing information such as age, gender, ethnicity, employment and housing. For the purposes of this experiment a subset of the overall dataset was selected; the geographical region of the West Midlands  (containing 104,267 records, 9.34\% of the total dataset). Twelve variables were selected, 1 numerical and 11 categorical, described in Appendix A. Minimal pre-processing was applied, and missing values were retained. 

\subsection{System selection}
The statistical methods used were Synthpop, developed by~\textcite{Nowok2016Synthpop:R} and DataSynthesizer, developed by~\textcite{Ping2017DataSynthesizer:Datasets}. The GAN methods were CTGAN, developed by~\textcite{Xu2019ModelingGAN} and TableGAN, developed by~\textcite{Park2018DataNetworks}. All methods were used with default parameters. It is recognised that the default parameters may not always produce the optimal performance (particularly with GANs) but the defaults are those most commonly applied and are used to provide a fair comparison across all methods.

Synthpop, an open source package written in R, by default uses methods based on classification and regression trees (CART, developed by~\textcite{Breiman1984ClassificationTrees}), which can handle mixed data types and is non-parametric. Synthpop synthesises the data sequentially, one variable at a time; the first is sampled, then the following are predicted using CART (in the default mode) with the previous variables used as predictors. This means that the order of variables is important (and can be set by the user). As suggested by~\textcite{Raab2017GuidelinesData}, variables with many categories may be moved to the end of the sequence, therefore the ordering was set by the least to maximum number of categories, with age first. \textcite{Raab2017GuidelinesData} also suggest that data rules should be included if they exist; since the SARs data contained four rules (e.g. if age $\leq$ 15, then marital status is single) these were included.

DataSynthesizer, a Python package, implements a version of the PrivBayes~\parencite{Zhang2017PrivBayes:Networks} algorithm. DataSynthesizer learns a differentially private Bayesian Network which captures the correlation structure between attributes and then draws samples. A settable parameter, $\epsilon$, controls differential privacy; 1 was used as the default. For the selected GAN based methods, CTGAN (described in the previous section) is a Python package developed to deal with mixed data; default parameters were used. TableGAN, implemented in Python, has a low, medium and high privacy setting; low was used as the default.


\subsection{Measuring Disclosure Risk using TCAP}

\textcite{Taub2018DifferentialExploration} introduced a measure for disclosure risk of synthetic data called the Correct Attribution Probability (CAP). We will be using an adaptation used in \textcite{Taub2019challenge} called the \textit{Targeted Correct Attribution Probability} (TCAP). The TCAP method is based on a strong intruder scenario in which two data owners produce a linked dataset (using a trusted third party), which is then synthesised and the synthetic data published. The adversary is one of the data owners who attempts to use the synthetic data to make inferences about the others' dataset. 

More modestly, at the individual record level, the adversary is somebody who has partial knowledge about a particular population unit (including the values for some of the variables in the dataset -- the keys -- and knowledge that the population unit is in the original dataset) and wishes to infer the value of a sensitive variable (the target) for that population unit. 

Following~\textcite{Taub2019challenge}, we assume that the adversary will focus on records which are in equivalence classes with corresponding $l$-diversity of 1 on the target and attempts to match them to their data. The TCAP metric is then the probability that those matched records yield a correct value for the target variable (i.e. that the adversary makes a correct attribution inference). 

Following~\textcite{Taub2019challenge}, TCAP is calculated as follows:
We define $d_{o}$ as the original data, and $K_{o}$ and $T_{o}$ as vectors for the key and target information, respectively
\begin{equation}
  d_{o}= \{K_{o} , T_{o}\}\quad.
\end{equation}
Likewise, $d_{s}$ is the synthetic dataset 
\begin{equation}
    d_{s}= \{K_{s} , T_{s}\}\quad.
\end{equation}

\noindent We then calculate the Within Equivalence Class Attribution Probability (WEAP) score for the synthetic  dataset. The WEAP score for the record indexed \textit{j} is the empirical probability of its target variables given its key variables

 \begin{equation}
\mathit{WEAP}_{s,j}=\mathit{Pr}(T_{s,j}|K_{s,j})=\frac{\sum
_{i=1}^{n}[T_{s,i}=T_{s,j},K_{s,i}=K_{s,j}]}{\sum
_{i=1}^{n}[K_{s,i}=K_{s,j}]}\quad,
\end{equation}   

\noindent where the square brackets are Iverson brackets, $n$ is the number of records, and $K$ and $T$ are vectors for the key and target information, respectively. Then using the WEAP score the synthetic dataset will be reduced to records with a WEAP score that is 1. 

\noindent The TCAP for record \textit{j} based on a corresponding original dataset
\textit{d}\textit{\textsubscript{o}} is the same empirical, conditional
probability but derived from \textit{d}\textit{\textsubscript{o}},

\begin{equation}
\mathit{TCAP}_{o,j}=\mathit{Pr}(T_{s,j}|K_{s,j})_{o}=\frac{\sum
_{i=1}^{n}[T_{o,i}=T_{s,j},K_{o,i}=K_{s,j}]}{\sum
_{i=1}^{n}[K_{o,i}=K_{s,j}]}\quad.
\end{equation}

\noindent For any record in the synthetic dataset for which there is no corresponding record in the original dataset with the same key variable values, the denominator in Equation 4 will be zero and the TCAP is therefore undefined. If the TCAP score is 0 then the synthetic dataset carries little risk of disclosure; if the dataset has a TCAP score close to 1, then for most of the riskier records disclosure is possible.

\subsection{Evaluating utility}

Following~\textcite{Taub2020TheRecords}, we assess the utility of the synthetic data using three measures: confidence interval overlaps (CIO), the ratio of estimates (ROE) and the propensity mean squared error (pMSE). To calculate the CIO we use 95\% confidence intervals, and use the coefficients from regression models built on the original and synthetic datasets. The CIO, proposed by~\textcite{Karr2006FrameworkConf}, is defined as:
\begin{equation}J_{k}= \frac{1}{2}(\frac{U_{,k}-L_{,k}}{U_{orig,k}-L_{orig,k}}+\frac{U_{,k}-L_{,k}}{U_{syn,k}-L_{syn,k}})\quad, \end{equation}
where $U_{,k}$ and $L_{,k}$ denote the respective upper and lower bounds of the intersection of the confidence intervals from both the original and synthetic data for estimate $k$, $U_{orig,k}$ and $L_{orig,k}$ represent the upper and lower bounds of the original data, and $U_{syn,k}$ and $L_{syn,k}$ of the synthetic data. 

ROE is calculated by taking the ratio of the synthetic and original data estimates, where the smaller of these two estimates is divided by the larger one. Thus, given two corresponding estimates (e.g. totals, proportions), where $y_{orig}^{1}$ is the estimate from the original data and  $y_{synth}^{1}$ is the corresponding estimate from the synthetic data, the ROE is calculated as:
\begin{equation} ROE = \frac{\mbox{min}(y_{orig}^{1}, y_{synth}^{1})}{\mbox{max}(y_{orig}^{1}, y_{synth}^{1})}\end{equation}
If $y_{orig}^{1} = y_{synth}^{1}$ then the ROE = 1. The ROE will be calculated over bivariate and univariate data, and takes a value between 0 and 1. For each categorical variable the ratio of estimates are averaged across categories to give an overall ratio of estimates.

The propensity score, or pMSE, developed in~\textcite{Woo2009GlobalLimitation} and~\textcite{Snoke2018GeneralData} is a measure of data utility designed to determine how easy it is to discern between two datasets based upon a classifier. It is calculated by merging the original and synthetic datasets and creating a variable $T$, where $T=1$ for the synthetic dataset and $T=0$ for the original dataset. For each record in the combined dataset the probability of being in the synthetic dataset is computed; this is the propensity score. The propensity score can be computed via logistic regression. The distributions of the propensity scores for the original and synthetic data are compared; where these are similar data utility should be high. In summary:
\begin{equation}
    pMSE = \frac{1}{N}\sum_{i=1}^{N}[\hat{p}_i - c]^2\quad,
\end{equation}
where $N$ is the number of records in the merged dataset, $\hat{p}_i$ is the estimated propensity score for record $i$, and $c$ is the proportion of data in the merged dataset that is synthetic (which is often $1/2$). A pMSE score close to 0 would indicate high utility (a score of 0 indicates the original and synthetic data are identical). 

\section {Results}
\label{sec:Results}

Synthetic datasets the same size as the original ($n=104,267$) were generated (aside from TableGAN which generated $n=104,000$ records). No post-processing of the data was performed. Figure \ref{fig:Comparison} in Appendix B shows indicative bar graphs for four of the variables. These plots illustrate that whilst the data produced by Synthpop and DataSynthesizer (and to a lesser extent CTGAN) show similar counts to the original dataset, the data produced by TableGAN in many cases does not. TableGAN did not identify some categories (for example, classifying all individuals as having zero higher educational qualifications, or no individual as being divorced) or over/under estimated in other cases (e.g. classifying a large proportion of individuals with missing Tenure).

\subsection{Disclosure risk}

Table 1 shows the TCAP score for each key, with LTILL (long-term illness), FAMTYPE (family type) and ECONPRIM (primary economic position) as the targets. Four sets of keys, ranging from three to six variables were used (see Appendix A).

\subsection{Utility}
Table 2 presents all utility metrics.  Following~\textcite{Taub2019challenge}, $1 - 4pMSE$ is used to scale the pMSE between 0 and 1. The mean of the ROE scores for the univariate and bivariate cross-tabulations (applied across all 66 possible combinations of two variables) is also presented. The CIO and standardized difference are presented as the mean across multiple regression models using each variable in the dataset as a target. The CIO could not be calculated for TableGAN because the variables in the dataset had insufficient variation to provide comparable logistic regression models.

The overall utility score is calculated by taking the mean of the ROE scores, the CI overlap and the value for $1 - 4pMSE$.  The overall utility score for TableGAN takes the mean excluding the CIO, which could not be calculated. The R-U (Risk-Utility) confidentiality map shown in Figure \ref{fig:RUmap} plots the overall utility score against the average TCAP (risk) score. Included for reference is the original data (which has an overall utility and risk score of 1) and the average TCAP baseline.

\begin{center}
  \small
  \begin{tabular}{ccccccc}
    \toprule
    Target & Key & Synthpop & DataSynthesizer & CTGAN & TableGAN & Baseline  \\
    \midrule
    LTILL   & 6 & 0.935 & 0.929 & 0.912 & 0.911 \\
            & 5 & 0.897 & 0.898 & 0.891 & 0.907 & 0.774 \\
            & 4 & 0.894 & 0.899 & 0.889 & 0.907 \\
            & 3 & 0.936 & 0.951 & 0.931 & 0.901 \\
    \addlinespace
    FAMTYPE  & 6 & 0.709 & 0.23 & 0.598 & 0.301 \\
             & 5 & 0.725 & 0.658 & 0.639 & 0.384 & 0.223 \\
             & 4 & 0.736 & 0.654 & 0.651 & 0.416 \\
             & 3 & 0.809 & 0.608 & 0.648 & 0.420 \\
    \addlinespace
    TENURE   & 6 & 0.596 & 0.429 & 0.490 & 0.217 \\
             & 5 & 0.504 & 0.376 & 0.453 & 0.336 & 0.329 \\
             & 4 & 0.500 & 0.350 & 0.447 & 0.341 \\
             & 3 & 0.496 & 0.353 & 0.482 & 0.279 \\
      \midrule
    Average  &   & 0.728 & 0.644 & 0.669 & 0.527 & 0.442\\
     \bottomrule
  \end{tabular}\\
  Table 1: TCAP scores.
\end{center}

\section{Discussion}
\label{sec:discussion}
The results show that the orthodox statistical method -- CART using Synthpop -- shows the highest utility and highest risk of the four methods tested (for the data and the parameter settings used). Table GAN produced the lowest level of risk but at levels of data quality that appeared to be unacceptably low. The other two methods were between the two.

It should be noted that this study is a proof of concept for the methodology and we make no assertions at this stage about generalisability. The limitations are:
\begin {itemize}
\itemsep0em
\item We have only tested the methods on a single dataset.
\item We have only tested here a sub-sample of records. Ideally we would want these to be applied to a population file.
\item We have used the default settings for each of the systems. Different settings would produce different outcomes.
\item We have only considered a small selection of GANs and no other machine learning methods. 
\item At present, the methods we are employing here are only attempting to optimize the closeness of the synthetic data to the original (i.e. analog utility). Ideally we would want a system that optimises risk and utility. 
\end{itemize}

\begin{figure}[h]
\centering
\includegraphics[scale=0.85]{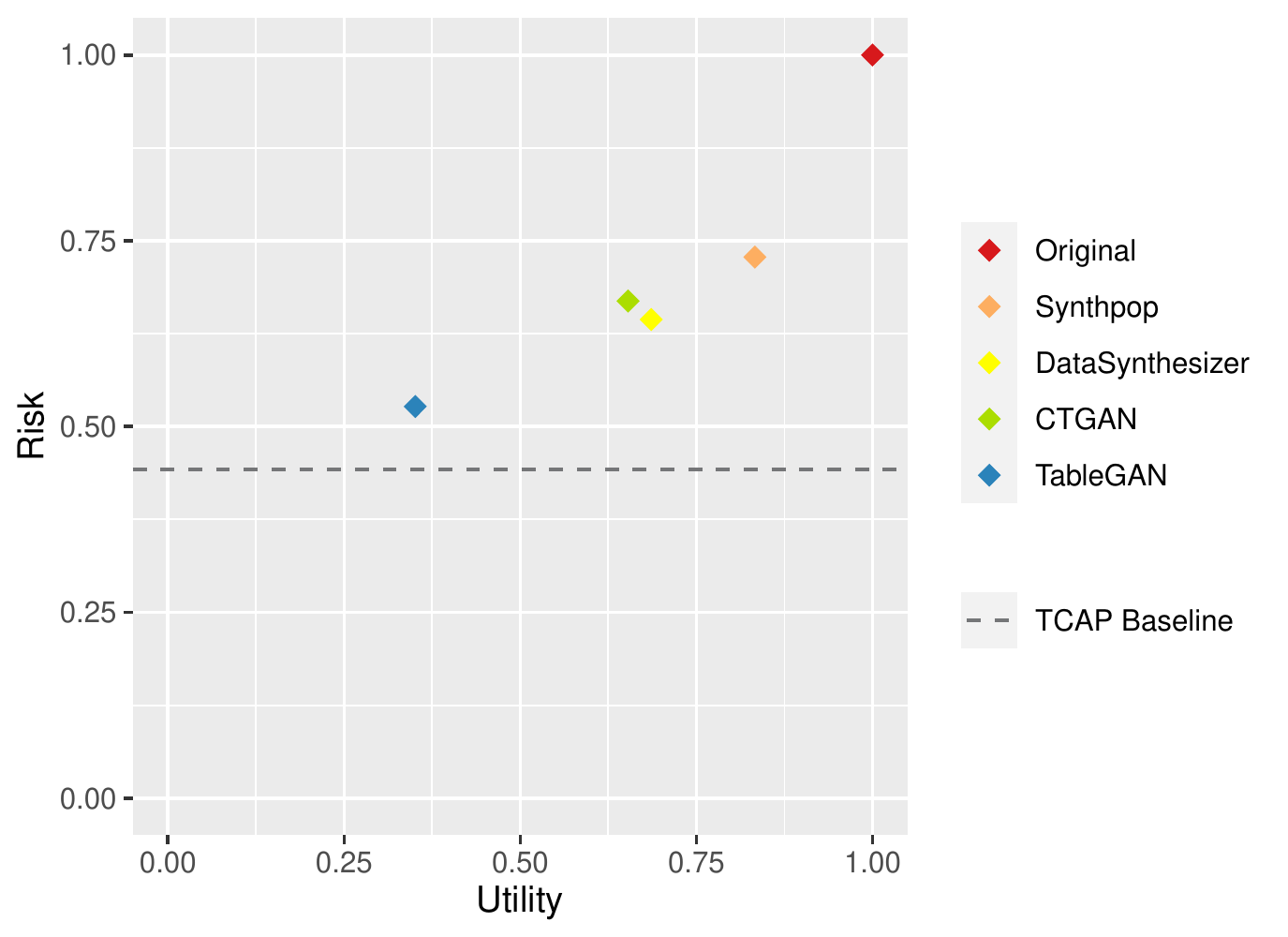}
\caption{\label{fig:RUmap}The Risk-Utility map of the synthetic datasets.}
\end{figure}

Notwithstanding the above limitations, Figure~\ref{fig:RUmap} does indicate that the trade off is happening even with synthetic data. The four data points (excluding the original data) that we have appear to fall on a straight line, and with the summary measure we have used the utility is traded quite heavily for reductions in risk. 
\begin{center}
  \begin{tabular}{ccccc}
    \toprule
    Metric & Synthpop & DataSynthesizer & CTGAN & TableGAN \\
    \midrule
    pMSE               & \textbf{0.00015} & 0.01438  & 0.03162  & 0.17529 \\
    log(pMSE ratio)    & \textbf{1.058}   & 5.655    & 6.442    & 8.153 \\
    1-(4 x pMSE)       & \textbf{0.9994}  & 0.9425   & 0.8735   & 0.2988 \\
   \addlinespace
    ROE univariate     & \textbf{0.981}   & 0.821 & 0.743 & 0.499 \\
    ROE bivariate      & \textbf{0.847}   & 0.616 & 0.587 & 0.255 \\
    \addlinespace
    CI Overlap         & \textbf{0.506}      & 0.365 & 0.410 & -\\
    Standardized difference & \textbf{3.721} & 4.297 & 4.034 & -\\
    \midrule
    Overall utility    & \textbf{0.833}   & 0.686 & 0.653 & 0.351 \\ 
   \bottomrule
  \end{tabular}\\
  Table 2: pMSE, ROE and CIO scores with best results for each measure in \textbf{bold}.
\end{center}

As a general observation on the methodology, the TCAP measure is itself relatively new and has never been calibrated. TCAP is a natural evolution of DCAP~\parencite{Taub2018DifferentialExploration}
(which can be regarded as an absolute but unrealistic measure of inferential disclosure risk). TCAP poses a more realistic intruder scenario but ideally we would want to pen test this scenario using a methodology such as that of~\textcite{elliott2016end}. 
In future work we will:
\begin{enumerate}
\itemsep0em
\item Run a much wider range of tests examining the effects of changes in parameter settings on each methods position on the RU map.
\item Investigate other GAN architectures with a view to developing the most appropriate one for the data synthesis task.
\item Test the above on larger datasets in terms of the number of variables and cases and investigate the extent to which it is possible to synthesise a whole (UK) census. 
\end{enumerate}

\printbibliography{}

\pagebreak
\begin{appendices}


\section{The SARS Dataset variables}
Twelve variables from the 1991 Great Britain Individual Sample of Anonymised Records~\parencite{OfficeforNationalStatisticsCensusDivisionUniversityofManchesterCathieMarshCentreforCensusandSurveyResearch2013CensusSARs}, or SARs dataset were used:
\begin{itemize}
    \item AREAP: individual SAR area, 21 categories
    \item AGE: age, integer range 0-95
    \item COBIRTH: country of birth, 13 categories
    \item ECONPRIM: primary economic position, 10 categories
    \item ETHGROUP: ethnic group, 10 categories
    \item FAMTYPE: family type, 9 categories
    \item LTILL: limiting long-term illness, 2 categories
    \item MSTAUS: marital status, 5 categories
    \item QUALNUM: number of higher educational qualifications, 3 categories
    \item SEX: sex, 2 categories
    \item SOCLASS: social class, 9 categories
    \item TENURE: tenure of household space, 7 categories
\end{itemize}

Note, the COBIRTH variable was aggregated to 13 categories (countries of the UK, Ireland and continents) as Synthpop can struggle with datasets containing many categorical attributes with many categories.

\subsection{Key variables used for TCAP}
For each of the target variables (LTILL, FAMTYPE and TENURE) the following were used as key variables:

\begin{itemize}
\itemsep0em
  \item 6 keys: AREAP, AGE, SEX, MSTATUS, ETHGROUP, ECONPRIM
  \item 5 keys: AREAP, AGE, SEX, MSTATUS, ETHGROUP
  \item 4 keys: AREAP, AGE, SEX, MSTATUS
  \item 3 keys: AREAP, AGE, SEX
\end{itemize}
\pagebreak
\section{Indicative plots of univariates}

\begin{figure}[h]
\centering
\includegraphics[width=1\textwidth]{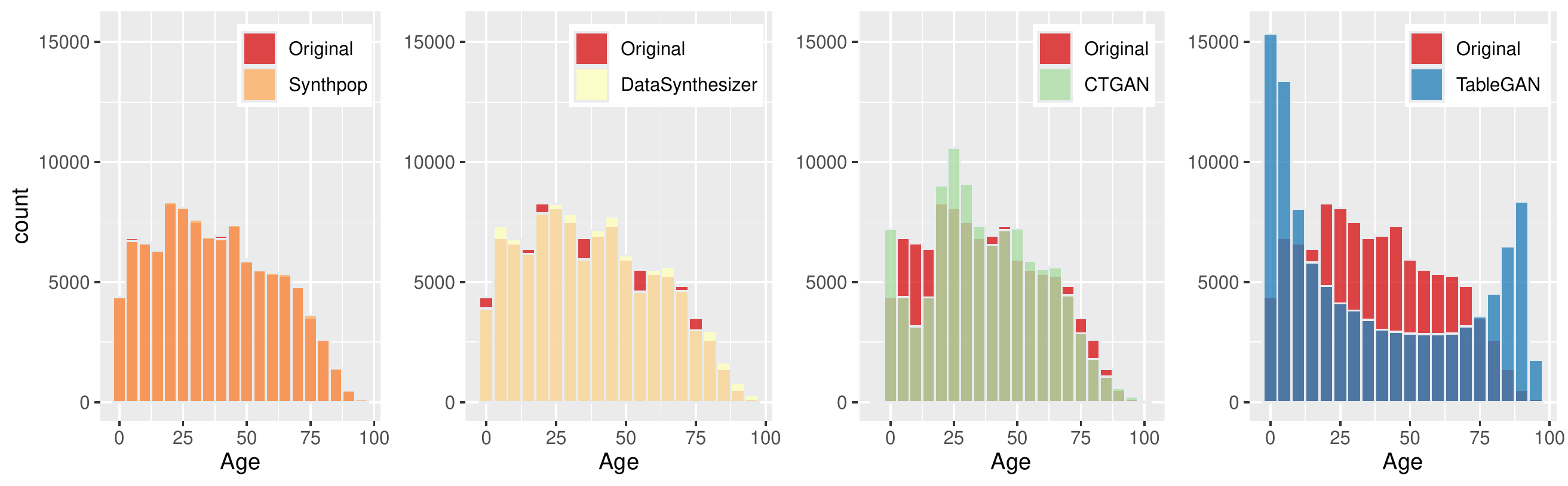}
\setlength{\abovecaptionskip}{-5pt}
\setlength{\belowcaptionskip}{-65pt}
\caption{\label{fig:Age}Histograms comparing original data with synthetic data for age}
\end{figure}

\vspace{3cm}

\begin{figure}[h]
\centering
\includegraphics[scale=0.7]{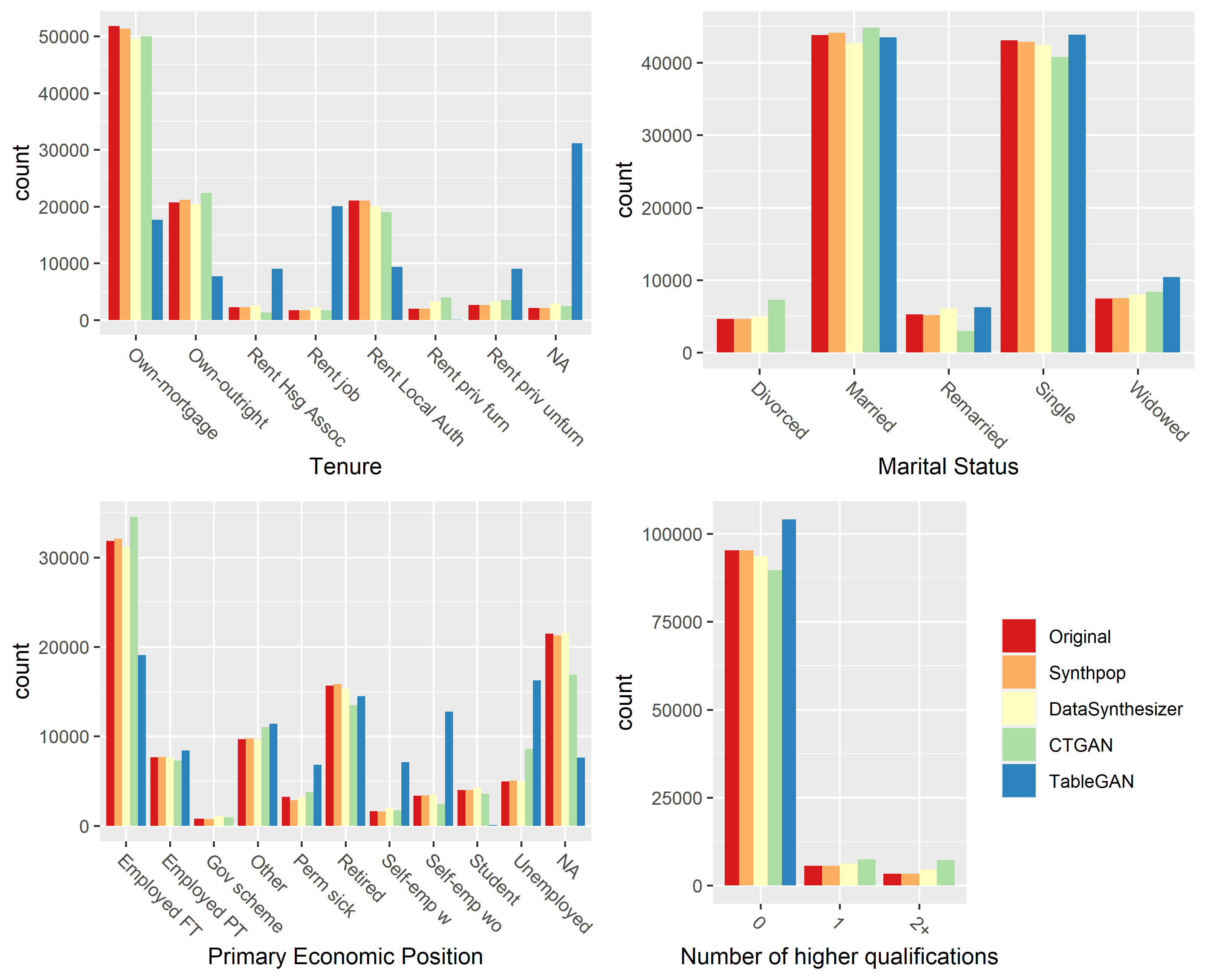}
\setlength{\abovecaptionskip}{-5pt}
\caption{\label{fig:Comparison}Bar graphs comparing original data to synthetic data}
\end{figure}
\end{appendices}

\end{document}